\documentclass[letterpaper, 10 pt, conference]{ieeeconf}  
\usepackage{lipsum} 
\usepackage{pifont}
\usepackage{amsmath}
\usepackage{amssymb}
\usepackage{mathtools}
\usepackage{stfloats}
\usepackage{subcaption}
\usepackage{float}
\usepackage{tabularx}
\usepackage{booktabs}

\IEEEoverridecommandlockouts                              

\title{\LARGE \bf
No More Marching: Learning Humanoid Locomotion \\for Short-Range SE(2) Targets
}

\author{Pranay Dugar$^{1}$, Mohitvishnu S. Gadde$^{1}$, Jonah Siekmann$^{2}$, Yesh Godse$^{2}$, Aayam Shrestha$^{1}$, Alan Fern$^{1}$ %
\thanks{*This work is supported by NSF Award 2321851, DARPA contract HR0011-24-9-0423, and the NVIDIA Academic Grant Program.} %
\thanks{$^{1}$Collaborative Robotics and Intelligent Systems Institute, Oregon State University, Corvallis, OR 97331. }%
\thanks{\{dugarp, gaddem, aayam.shrestha, alan.fern\}@oregonstate.edu} %
\thanks{$^{2}$Agility Robotics, Salem, OR 97317.}%
\thanks{\{jonah.siekmann, yesh.godse\}@agilityrobotics.com} %
}
\begin{document}

\maketitle
\thispagestyle{empty}
\pagestyle{empty}


\begin{abstract}
Humanoids operating in real-world workspaces must frequently execute task-driven, short-range movements to SE(2) target poses.
To be practical, these transitions must be fast, robust, and energy efficient. 
While learning-based locomotion has made significant progress, most existing methods optimize for velocity-tracking rather than direct pose reaching, resulting in inefficient, marching-style behavior when applied to short-range tasks. 
In this work, we develop a reinforcement learning approach that directly optimizes humanoid locomotion for SE(2) targets.
Central to this approach is a new constellation-based reward function that encourages natural and efficient target-oriented movement. 
To evaluate performance, we introduce a benchmarking framework that measures energy consumption, time-to-target, and footstep count on a distribution of SE(2) goals. 
Our results show that the proposed approach consistently outperforms standard methods and enables successful transfer from simulation to hardware, highlighting the importance of targeted reward design for practical short-range humanoid locomotion.
\end{abstract}

\section{INTRODUCTION}

One of the defining capabilities of humanoid robots is legged locomotion, which can be broadly categorized into two functional modes. \emph{Long-range locomotion} involves generating stable, rhythmic gaits guided by velocity commands, and is well suited for traversing extended distances. In contrast, \emph{short-range locomotion} focuses on task-driven movements toward specific $\mathrm{SE}(2)$ target poses, and dominates many practical workflows---for example, when a robot repeatedly moves boxes between nearby locations. This second mode of locomotion differs fundamentally from gaited walking: it prioritizes rapid, adaptive transitions and functional precision over rhythmic footstep generation. Consequently, applying long-range locomotion controllers to short-range $\mathrm{SE}(2)$ targets will often produce inefficient and unnatural marching-style behaviors.

Reinforcement learning (RL) has recently enabled state-of-the-art performance in long-range, velocity-target humanoid locomotion \cite{RealHumanoid2023, zhang2024achieving, Marum2024RevisitingRD, li2025reinforcement}. These successes have largely hinged on well-designed reward functions that promote forward velocity, gait stability, and symmetry—principles validated across a range of humanoid and quadruped platforms. However, such reward structures are not directly applicable to short-range $\mathrm{SE}(2)$-target locomotion, where the robot must simultaneously make progress toward a target position and align to a desired orientation. This dual-objective introduces ambiguity: should the robot rotate in place before stepping forward, reach the position first and then turn, or rotate while advancing? Naive reward formulations that resolve this ambiguity too rigidly or too loosely can produce behavior that is inefficient, unnatural, and poorly suited to the demands of confined, task-driven environments. These challenges highlight the need for reward functions explicitly tailored to $\mathrm{SE}(2)$-target commands.

\begin{figure}[t]
  \centering
  \includegraphics[width=0.85\columnwidth]{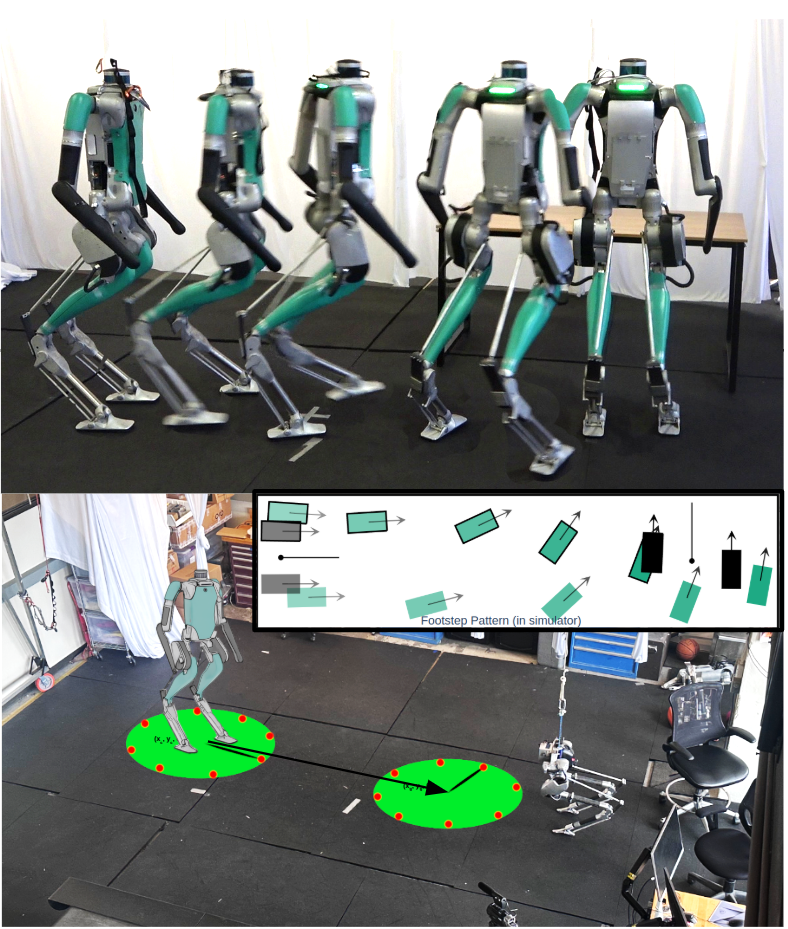}
  
  \caption{\small
  Overview of our approach for short-range $\mathrm{SE}(2)$-target locomotion.  
  \textbf{Top:} The learned \textit{GoTo} controller produces coordinated motion toward a specified goal pose.  
  \textbf{Middle:} Footstep pattern generated by the GoTo policy for reaching a commanded $\mathrm{SE}(2)$ target.  
  \textbf{Bottom:} Real-world deployment of the controller on the Digit humanoid, walking from an initial pose $(x_c, y_c, \theta_c)$ to a goal pose $(x_g, y_g, \theta_g)$. The green circle and dots denote the constellation used to define the reward, which aligns translation and rotation simultaneously, enabling efficient, natural locomotion without marching-style behavior.
  }
  \label{figure:main_img}
  \vspace{-1em}
\end{figure}

In this paper, we explore end-to-end reinforcement learning for short-range $\mathrm{SE}(2)$-target locomotion for humanoid robots. Our main contribution is the point constellation reward---a minimally constraining objective that promotes efficient and natural motion. Inspired by 
loss functions from 
learning-based motion planning for fixed-base manipulators \cite{fishman2023, allshire2022transferring, Portela2024WholebodyEP}, the reward is based on the geometric similarity between point clouds representing the robot's current and target configurations. We adapt this idea to legged $\mathrm{SE}(2)$-target locomotion by defining a virtual 2D constellation of points rigidly attached to the robot's base, along with a corresponding constellation at the target pose. At each timestep, the reward is computed from the alignment between the current and target constellations, yielding a unified geometric signal that captures both translation and rotation. We show that the trade-off between positional and rotational accuracy is governed by the moment of inertia of the constellation, offering intuitive control and flexibility across tasks and workspace layouts.

Using this reward, we train and evaluate both end-to-end policies and hierarchical architectures built on top of a pre-trained velocity-based controller, and show that even simple policy structures can produce robust, efficient behaviors. To assess effectiveness, we measure key aspects of locomotion performance, including final pose error, energy consumption, time to target, and footstep count. Our simulation results demonstrate that the constellation-based reward combined with end-to-end policy training leads to significantly more natural and efficient movement compared to learned or hand-coded approaches based on velocity-tracking. Moreover, the formulation reveals meaningful trade-offs between translational and rotational accuracy, allowing fine-grained behavioral control through reward shaping. Finally, we demonstrate successful sim-to-real transfer on digit-v3 humanoid platform confirming the method’s practical viability.

The key contributions of this work are: (1) an end-to-end RL approach with a constellation-based reward that intuitively balances translational and rotational objectives for short-range $\mathrm{SE}(2)$-target locomotion; (2) a rigorous benchmarking framework $\mathrm{SE}(2)$-target locomotion that quantifies locomotion performance using interpretable metrics, and (3) experimental validation in both simulation and hardware, demonstrating improved efficiency, robustness, and real-world applicability in constrained environments.

\section{RELATED WORK}

\subsection{Model-Based Control for Humanoid Navigation}
A common approach to humanoid navigation is to decompose the problem into global and local planning. Systems such as navfn \cite{macenski2022robot} are used for high-level path planning, while local planners translate segments of these paths into executable footstep sequences. The humanoid\_planner\_2d package exemplifies this architecture by leveraging SBPL algorithms for 2D planning, integrated with the footstep\_planner module to generate feasible foot placements.

Several works have extended this paradigm to more complex environments. For example, Garimort et al. \cite{garimort2011humanoid} developed methods for selecting optimal footstep sequences in cluttered spaces, allowing humanoids to navigate around or over obstacles. The Open Source Step-Informed framework (OpenSIn) \cite{Stumpf2016OpenSI} adds full 3D footstep planning integrated with perception and locomotion systems. More recently, Mishra et al. \cite{Mishra2022FootstepPO} proposed using generative adversarial networks (GANs) to generate footstep plans for navigation in highly constrained indoor environments, such as narrow corridors or cluttered rooms.

These footstep planning systems typically take start and goal poses (including orientation) and output a discrete sequence of footsteps to reach the target. Replanning is supported to adapt to dynamic obstacles or localization drift. However, executing these plans requires an additional layer of complexity: (1) an inverse kinematics solver to compute joint angles for foot placement, (2) a trajectory generator for smooth transitions between poses, and (3) a low-level controller to track these trajectories while maintaining balance.

In contrast, we focus on end-to-end learning of locomotion policies for point-to-point navigation in constrained workspaces. By designing an expressive reward function tailored to this task, we eliminate the need for explicit footstep planning or trajectory stitching, enabling more adaptive and efficient control with reduced system overhead.

\subsection{Learning-Based Control for Humanoid Navigation}

Significant work has been done on learning-based approaches to basic humanoid locomotion. Early work \cite{siekmann2020learning, siekmann2021blind} used reinforcement learning (RL) for sim-to-real transfer focusing on periodic gait patterns. More recently, Li et al. \cite{li2024reinforcement} extended this approach to support both periodic and non-periodic motions within a unified framework. While other works \cite{Zhang2025NaturalHR,Ma2025StyleLocoGA} use techniques such as generative motion priors to produce more human-like walking. However, these methods primarily emphasize forward, velocity-target locomotion and do not address the efficiency or control challenges of short-range $\mathrm{SE}(2)$-target navigation. 

To address higher-level navigation tasks, previous learning-based approaches have typically adopted  a hierarchical control architecture, where a high-level policy predicts target velocities, and a low-level controller tracks these commands using standard walking gaits. For example, Kenzo et al. \cite{lobosTsunekawa2018VisualNF} uses a vision-based navigation policy on top of an omnidirectional velocity controller to reach the target locations. Similarly, \"Ozaln et al. \cite{zaln2019AnIO} employs a discrete set of velocity commands as the low-level control interface. While effective in structured settings, these approaches inherently produce marching-style behaviors, which are suboptimal for short-range, task-oriented locomotion in constrained spaces.

Imitation-based frameworks for simulated humanoid characters have made significant progress in capturing the “naturalness” of human motion. These approaches typically train robust low-level controllers from motion capture data—either to implicitly follow commanded velocities \cite{shrestha2024generating}, or to learn a latent space of motion skills and control behavior through a hierarchical policy over that space \cite{Luo2023UniversalHM, ASE, CASE, CALMC}. However, these methods remain largely confined to simulation and have not yet been successfully transferred to real humanoid platforms. Moreover, most of these systems rely on Adversarial Motion Priors (AMP) \cite{AMP} to generate lifelike behaviors, which depend on clean motion datasets and a non-trivial retargeting pipeline. This requirement becomes especially challenging for robots with non-human morphology—for example, the Digit v3 platform, which includes closed kinematic loops and a backward-bent knee design.

In contrast, our work proposes a reward-driven, end-to-end learning framework that does not rely on predefined velocity primitives or motion priors. By focusing on reward design tailored to $\mathrm{SE}(2)$-target locomotion, we aim to produce behaviors that are better aligned with short-range functional tasks and more transferable to physical robots.

\section{Problem Formulation }
\label{sec:prob_form}

We consider the problem of enabling a bipedal humanoid robot to perform efficient short-range locomotion toward a specified $\mathrm{SE}(2)$ target pose. We formulate this as a goal-conditioned Markov decision process (MDP) defined by the tuple $(S, G, A, p, R, \gamma, \rho_0, \rho_g)$, where $S$ is the state space consisting of proprioceptive measurements $q_t$ (joint positions and velocities) and the robot's current $\mathrm{SE}(2)$ pose $p_r = (x_r, y_r, \theta_r)$ in the world frame, $G$ is the goal space of $\mathrm{SE}(2)$ target poses $p_g = (x_g, y_g, \theta_g)$, $A$ is the action space of joint position commands, $p(s_{t+1} | s_t, a_t)$ is the dynamics function. $\rho_0$ is the initial state distribution of robot, and $\rho_g$ is the goal distribution which consists of random $\mathrm{SE}(2)$ target poses around the workspace. 

Given this MDP framework, our objective is to learn a goal-conditioned policy $\pi(a_t | s_t, g)$ that generates efficient and natural locomotion behaviors for navigating between arbitrary $\mathrm{SE}(2)$ poses in the workspace. In particular, our work emphasizes short-ranged targets, for which standard marching-style locomotion can be highly sub-optimal.  Unlike traditional learning-based locomotion controllers that optimize for velocity-tracking, our formulation directly optimizes for pose-reaching performance. We refer to this optimized policy as the \emph{GoTo} controller. 

{\bf Metrics.} To enable standardized comparison of $\mathrm{SE}(2)$-target controllers, we propose the following metrics that measure the accuracy, efficiency, and qualitative aspects of a controller. In our experiments, these metrics will be evaluated across a range of initial states and goals drawn from distributions of interest. 
\begin{itemize}
    \item \textit{Metric 1: Position Error.} This metric measures the Euclidean distance between the robot's final position ($x_f, y_f)$ and the target position $(x_g, y_g)$. For each trial, we compute the final position error as $\sqrt{(x_f - x_g)^2 + (y_f - y_g)^2}$. This provides a direct measure of how precisely the controller can position the robot in the $(x,y)$ plane.
    \item \textit{Metric 2: Orientation Error.} This metric measures the absolute angular difference between the robot's final heading $\theta_f$ and the target heading $\theta_g$. We compute this as $\min(|\theta_f - \theta_g|, 2\pi - |\theta_f - \theta_g|)$ to account for the circular nature of orientation. This quantifies the controller's ability to achieve the desired orientation independently of positional accuracy.
    \item \textit{Metric 3: Time to Target.} This metric captures the total time required for the robot to reach the target pose within acceptable error thresholds. A trial is considered successful if both position error and orientation error fall below predefined thresholds (e.g., position error \(<\) 5 cm and orientation error \(<\) 0.1 rad) when the velocity of the agent is zero. This metric evaluates the speed at which the controller can accomplish the navigation task.
    \item \textit{Metric 4: Footstep Count.} We measure the number of distinct foot placements made by the robot during the trajectory to the target. This serves as a proxy for motion efficiency and stability, with fewer steps generally indicating more efficient movement planning. Each step is counted when a foot transitions from a swing phase to a stance phase with the foot making contact with the ground.
    \item \textit{Metric 5: Energy Efficiency.} We quantify the total mechanical energy expended during the locomotion task by computing the integral of joint power over time: $E = \int_0^T \sum_{i=1}^n \tau_i(t) \cdot \omega_i(t) \, dt$, where $\tau_i(t)$ represents the torque of joint $i$ at time $t$, $\omega_i(t)$ is the angular velocity of joint $i$, $n$ is the total number of actuated joints, and $T$ is the total task time. This metric is normalized by the distance traveled to compute energy per meter, providing a consistent comparison across different target distances. Lower energy consumption generally indicates more efficient motion and extends the operational runtime of the robot in real-world applications.
\end{itemize}

\section{GoTo Reward Formulation}
\label{sec:reward}

In Section~\ref{sec:training}, we describe our reinforcement learning (RL) approach for training the GoTo controller. A critical factor in the success of this---or any---RL-based locomotion system is the design of the reward function, which must effectively guide learning toward desirable and transferable behaviors. For the $\mathrm{SE}(2)$-target locomotion task, the reward should encourage the agent to reach the specified target pose efficiently, naturally, and in a manner that supports successful sim-to-real transfer. To meet these objectives, our reward at each timestep is composed of three terms:
\begin{equation*}
r = r_{\text{con}} + r_{\text{sty}} + r_{\text{reg}}
\end{equation*}
Here, \( r_{\text{con}} \) is the proposed \emph{constellation reward}, which promotes task progress toward $\mathrm{SE}(2)$ targets; \( r_{\text{sty}} \) is a \emph{style reward} that encourages natural bipedal gaits; and \( r_{\text{reg}} \) is a \emph{regularization reward} that promotes energy efficiency, motion smoothness, and hardware compatibility. We describe each of these components in detail below.

\subsection{Constellation Reward Component} 

A fundamental question in $\mathrm{SE}(2)$-target locomotion is how to design a reward component that effectively encourages task completion, i.e. reaching the desired $\mathrm{SE}(2)$ pose. The most direct approach is to assign a reward of 1 when the robot is within a specified tolerance of the target, and 0 otherwise. However, such sparse rewards are impractical in high-dimensional, continuous control problems, where the likelihood of reaching the goal through exploration alone is vanishingly small. To address this, it is common to use denser rewards that provide a continuous ``progress signal” toward the goal. In our setting, this naturally suggests a distance measure that quantifies how close the current robot pose is to the target pose.

A natural approach is to treat translation and rotation as independent components and design separate reward terms for each. While conceptually straightforward, we found this structure extremely difficult to tune in practice. Small changes to the internal or relative weightings often led to unbalanced behavior in which one component dominates---for example, policies might consistently reduce position error before addressing orientation, or vice versa---resulting in unnatural ``stop-and-turn'' motion. Similar behaviors have been observed in prior work on humanoid animation and skill embedding~\cite{ASE, Luo2023UniversalHM}, where independently weighted $\mathrm{SE}(2)$ rewards are used for high-level control. However, those methods do not aim to produce efficient or natural $\mathrm{SE}(2)$-target transitions.

To overcome this challenge, we introduce the \emph{constellation reward}---a unified, interpretable geometric objective that jointly captures both translational and rotational alignment. This approach is inspired by point-set registration methods in computer vision~\cite{besl1992method, chen1992object, arun1987least}, where both translation and rotation are implicitly encoded through the alignment of two point constellations. Specifically, we represent the robot's $\mathrm{SE}(2)$ pose \((x_r, y_r, \theta_r)\) using a set of \(N\) landmark points \((p_1, \ldots, p_N)\) rigidly anchored to the robot's base frame. A change in pose corresponds to a rigid transformation of this point set. The target pose \((x_g, y_g, \theta_g)\) is similarly represented by the corresponding transformed constellation \((p_1^*, \ldots, p_N^*)\). The constellation distance is then defined as the mean squared distance between corresponding points:
\begin{equation}
d_{\text{con}} \;=\; \frac{1}{N} \sum_{i=1}^{N} \left\| p_i - p_i^{*} \right\|_2^2
\label{eq:constellation_loss}
\vspace{-0.15em}
\end{equation}
This distance provides a smooth signal that naturally couples position and orientation errors via the geometry of the constellation, without requiring hand-tuned relative weightings.

The constellation distance is zero when the robot is at the target pose, and increases with both positional and orientation error. The balance between these components is governed by the chosen constellation geometry. For example, constellations with small spatial extent relative to the robot emphasize positional accuracy, while larger constellations increase sensitivity to orientation. Importantly, we find that selecting a constellation that yields desirable behavior is far more straightforward than tuning separate position and orientation rewards. In practice, using a spatial extent comparable to the robot's natural stride length works well. 

To gain further insight into the trade-off between translation and rotation, we can derive an intuitive decomposition of the constellation distance. Let \( c = \tfrac{1}{N} \sum_i p_i \) and \( c^* = \tfrac{1}{N} \sum_i p_i^* \) be the centroids of the robot and target constellations, and define \( q_i = p_i - c \) and \( q_i^* = p_i^* - c^* \). The distance can then be decomposed as:
\begin{equation}
d_{\text{con}} = \underbrace{\lVert c - c^* \rVert_2^2}_{d_p}
+ \underbrace{\frac{1}{N} \sum_{i=1}^N \lVert q_i - q_i^* \rVert_2^2}_{\mathcal{L}_{\text{rot}}}
\end{equation}
Assuming the constellations differ only by a rotation \( R(\theta) \), i.e., \( q_i = R(\theta) q_i^* \), the rotational term becomes:

\begin{equation}
\mathcal{L}_{\text{rot}} = 2 I_c (1 - \cos\theta) \;\approx\; I_c \theta^2 \;\equiv\; I_c d_o
\end{equation}

where \( I_c = \tfrac{1}{N} \sum_i \lVert q_i \rVert_2^2 \) is the (constant) planar moment of inertia of the constellation. This yields a decomposed form:
\begin{equation}
d_{\text{con}} = d_p + I_c d_o
\label{eq:constellation_distance}
\end{equation}
which provides an intuitive geometric interpretation of the trade-off between translational and rotational accuracy, with \( I_c \) governing their relative influence.

To convert the constellation distance into a reward, we apply a negative exponential transformation:
\begin{equation}
r_{\text{con}} = e^{-w_c d_{\text{con}}} = e^{-w_c (d_p + I_c d_o)} = e^{-w_c d_p} \cdot e^{-w_c I_c d_o}
\label{eq:constellation_reward}
\end{equation}
where \( w_c \) determines how sharply the reward decays with error. Notably, this formulation results in a multiplicative coupling between translation and rotation. This is in contrast to the additive structure used in prior work~\cite{ASE, Luo2023UniversalHM}:
\begin{equation}
r_{\text{add}} = a_p \cdot e^{-w_p d_p} + a_o \cdot e^{-w_o d_o}
\end{equation}
which requires careful tuning of the additional parameters to balance the two components. Rather, the constellation reward provides an elegant and geometrically grounded alternative that achieves this balance implicitly, with clear physical interpretability through the moment of inertia. This coupling encourages coordinated reduction in both position and orientation error, yielding more efficient and natural $\mathrm{SE}(2)$-target behavior.

\subsection{Style and Regularization Reward Components}

We adopt the style and regularization rewards directly from prior work on sim-to-real training for standing and walking controllers~\cite{Marum2024RevisitingRD}. Due to space limitations, we refer the reader to that work for full implementation details. The style reward promotes natural body posture and gait characteristics during both standing and walking phases. It includes components that encourage appropriate base height, foot airtime, foot orientation, foot placement, and arm positioning. The regularization reward encourages smooth joint trajectories and energy-efficient behavior by penalizing large action changes, high torque magnitudes, and excessive base accelerations. Together, these components play a key role in facilitating robust sim-to-real transfer.

\section{GoTo Architecture and Training}
\label{sec:training}

In this section we describe the neural network architecture of our GoTo controller followed by the sim-to-real RL framework used for training. 

\subsection{Controller Architecture}
We implement the GoTo controller as a (128, 128) two-layer long short-term memory (LSTM) recurrent neural network. The \textit{input} to the network at each time step consists of: (1) the current robot state [joint positions, joint velocities, root orientation, and root angular velocities], and (2) the delta commands $\Delta x$, $\Delta y$, $\Delta \theta$ representing the difference between the robot's current $\mathrm{SE}(2)$ pose $(x_c, y_c, \theta_c)$ and the target pose $(x_g, y_g, \theta_g)$. The output action of the controller is the joint-space PD setpoints for all 20 actuators. The GoTo controller operates at 50Hz with the underlying PD controllers running at 1000Hz.

\subsection{Simulation-Based Training}

We train the controller using the Proximal Policy Optimization (PPO) algorithm~\cite{schulman2017ppo}, extended with a mirror loss to encourage symmetry in policy behavior \cite{mittal2024symmetry}. All training is done in simulation using the NVIDIA IsaacLab physics simulator for the Agility Digit-V3 robot model. 

{\bf Episode Generation.} We follow a standard episodic RL training process. Each episode initializes 4096 robot instances on a flat plane in the nominal standing posture with randomized $\mathrm{SE}(2)$ positions and orientations. Small uniform perturbations are applied to all joint angles around the nominal configuration, establishing the starting pose distribution $\rho_0$.
For the $\mathrm{SE}(2)$-target distribution $\rho_g$, we sample target poses such that $\Delta x \in [-2, 2]$ m, $\Delta y \in [-1.5, 1.5]$ m, and $\Delta \theta \in [-\pi, \pi]$ rad. New goal commands are issued at the start of each episode and then resampled at random intervals between 4 and 8 seconds. This allows for the robot to learn to react to new target commands even when in a dynamic motion state. The goals are drawn from five categories with given probabilities: stand-in-place (0.1), straight translation (0.2), lateral translation (0.2), turn in place (0.2), and combined rotation and translation (0.3). An episode terminates if the robot's base height falls below 0.4m or if the elapsed time reaches 8 seconds.

{\bf Constellation Reward.} We trained the GoTo controller using a constellation reward based on a circular constellation around the robot with radius $r=1\text{m}$, noting that the corresponding moment of inertia is $I_c=r^2=1$. The scaling factor in the exponential transform was selected to be $w_c=0.2$ based on a limited parameter sweep. 

{\bf Domain Randomization.} To support sim-to-real transfer we employ extensive domain randomization during each training episode. Our domain randomization is similar to Dugar et al.~\cite{dugar2025mhc} and van Marum et al.~\cite{Marum2024RevisitingRD} which randomizes joint stiffness, body masses, center-of-mass locations, encoder noise, observation noise, and torque delays. Additionally, we randomize the PD gains of all actuated joints by 10\% to enhance sim-to-real transfer. We also randomize foot friction parameters in order to simulate varying friction with the contact surface.

\section{Experimental Setup and Results}

We evaluate our proposed GoTo controller against natural baselines on the Digit V3 humanoid in simulation and in the real-world. Below we first describe the baselines, followed by quantitative and qualitative simulation results, and finally our real-world experiments. 

\subsection{Baseline Controllers}
To quantifiably assess the benefits of end-to-end learning with the constellation reward compared to traditional approaches we compare the GoTo controller against three representative baseline approaches, depicted in Figure \ref{figure:baselines}, which encompass different control paradigms for short-range $\mathrm{SE}(2)$ navigation:

\begin{figure}[t]
  \centering
  \includegraphics[width=0.95\columnwidth]{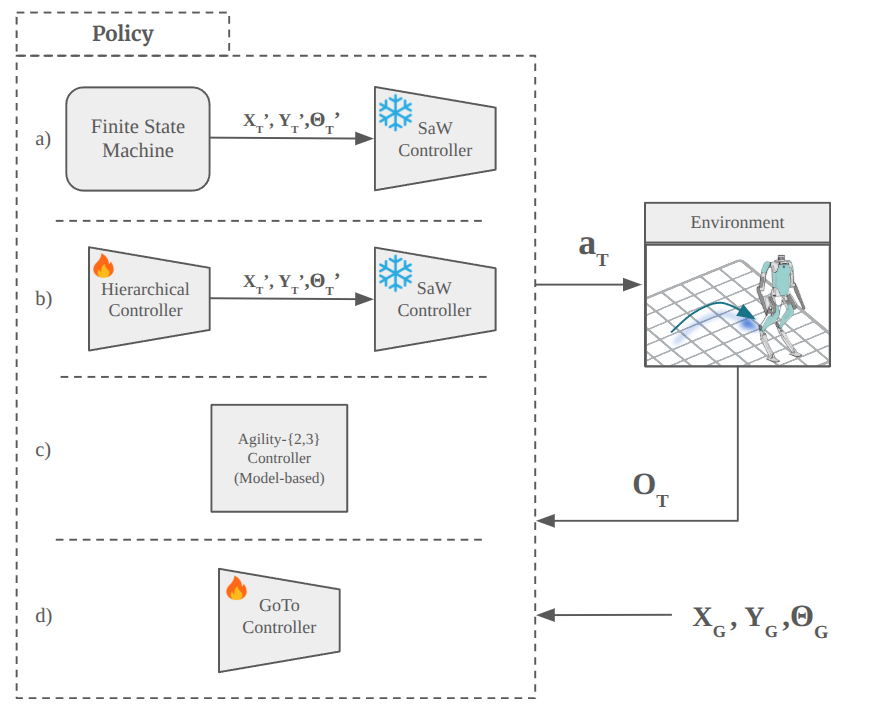}
  
  \caption{\small The figure shows three control architectures for humanoid locomotion to $\mathrm{SE}(2)$ targets. The Finite State Machine approach (a) uses a rule-based system to generate velocity commands for a Stand and Walk controller through a simple orient-then-move strategy. The Hierarchical Policy (b) employs a learned High-Level Controller that produces velocity commands for a pre-trained locomotion controller. The manufacturer provided controlelr (c) uses model based approach to reach target positions and can be tuned to use 2 or 3 phase strategy to reach the target.Our proposed GoTo Controller (c) implements an end-to-end learned approach that directly maps goal poses to actions without intermediate velocity commands. All systems interact with the environment by executing actions and receive observation and target $\mathrm{SE}(2)$ goal as input.
  }
  \label{figure:baselines}
  \vspace{-1em}
\end{figure}

\textit{Finite-State Machine (FSM):} A reactive strategy built on top of an existing velocity-controlled standing-and-walking controller ~\cite{Marum2024RevisitingRD}. The strategy first rotates the robot to face the target and then moves directly toward it using a trapezoidal velocity profile. The FSM embodies a classic "orient-then-move" approach and serves as a hand-tuned reference.

\textit{Agility Model-Based Controllers:} A manufacturer-supplied model-based footstep planner that optimizes footstep locations and orientations using an explicit dynamics model to minimize translation and heading errors. These are two different settings of this controller and we include both in our evaluation. \emph{Agility-3} is a 3 phase model where it first orients to the vector connecting the current position and target position, then moves to the position $(x, y)$, and then orients to the target heading. This is the default manufacturer controller, but it performs poorly for many $\mathrm{SE}(2)$ targets, e.g. when a single backward step would achieve the goal. \emph{Agility-2} is customized variant with 2 phases. It first orients to the target heading and then goes to the required $(x, y)$ position. 

\textit{Hierarchical Controller:} A learned high-level policy that takes $\mathrm{SE}(2)$ offsets as inputs and outputs reference base velocities for a pre-trained locomotion controller~\cite{Marum2024RevisitingRD} to track. This two-stage setup isolates the benefit of high-level learning on top a fixed gait generator as followed by previous works \cite{ASE, Luo2023UniversalHM}.

\begin{table*}[htbp]
    \centering

    \label{tab:your_label_here}
    {\fontsize{10}{12}\selectfont 
    \begin{tabular}{llllll}
\toprule
Method & Energy (KJ) ($\downarrow$) & Time Taken (s) ($\downarrow$) & Num footsteps ($\downarrow$) & Pos Error (cm) ($\downarrow$) & Theta error (rad) ($\downarrow$) \\
\midrule
Agility-3 & 2.9 ± 3.56 & 1.58 ± 0.86 & 1.69 ± 1.08 & 1.13 ± 0.64 & 1.77 ± 0.91 \\
FSM & 2.04 ± 0.91 & 1.35 ± 0.32 & 1.72 ± 0.51 & 1.33 ± 0.46 & 1.6 ± 0.77 \\
Hier. SaW & 1.11 ± 0.74 & 0.95 ± 0.28 & 1.05 ± 0.37 & 1.39 ± 0.61 & 1.63 ± 0.91 \\
Agility-2 & 1.0 ± 0.14 & 1.0 ± 0.12 & 1.0 ± 0.13 & 1.0 ± 0.6 & 1.0 ± 0.65 \\
\textbf{GoTo (Ours)} & \textbf{0.63 ± 0.19} & \textbf{0.56 ± 0.09} & \textbf{0.67 ± 0.15} & \textbf{0.6 ± 0.37} & \textbf{0.42 ± 0.31} \\
\bottomrule
\end{tabular}
}
    \caption{\small Comparison of our point-to-point GoTo controller with three baseline approaches: handcrafted Finite State Machine, factory-provided model-based controller (Agility), and Hierarchical Controller trained on velocity-based locomotion. Each metric is normalized using the corresponding value from hand tuned manufacturer-provided Agility Controller (Agility-2). The GoTo controller, learned using the proposed constellation reward, demonstrates superior efficiency with minimal energy usage and footsteps while maintaining competitive positional accuracy and the best orientation accuracy.}
    \label{tab:overall}

\end{table*}

\begin{figure*}[htbp]
  \centering
  \includegraphics[width=\linewidth]{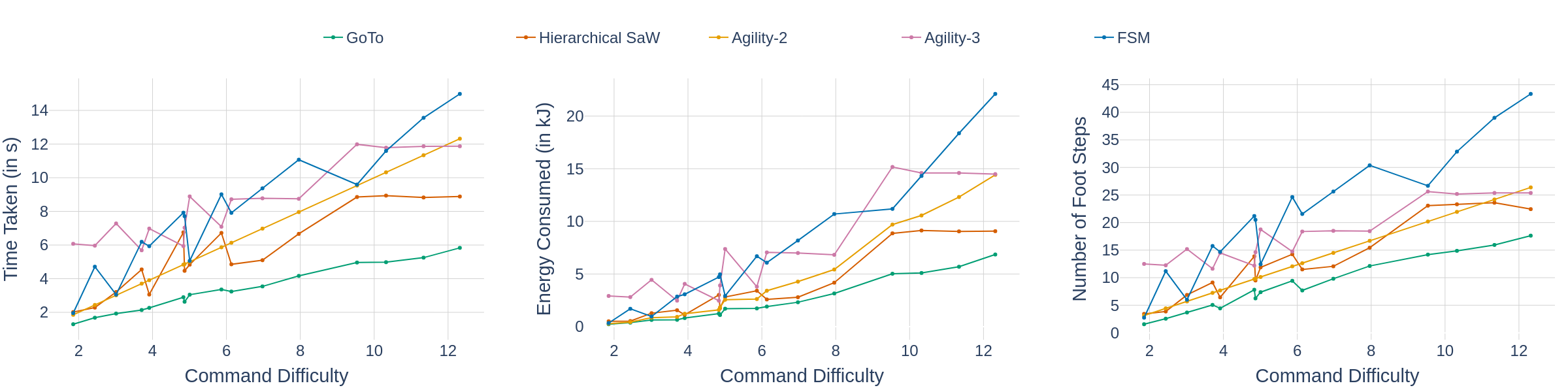}
  \caption{\small Controller performance across command difficulty. The difficulty metric (x-axis) represents the complexity of $\mathrm{SE}(2)$ target commands, measured by the number of timesteps Agility-2 requires to complete each distance-orientation combination. Results show time taken (left), energy consumed (center), and footstep count (right) for all controllers. The GoTo controller (green) maintains consistent efficiency advantages across all difficulty levels, with its benefits becoming increasingly pronounced for more complex commands that combine longer distances with larger orientation changes.}
  \label{fig:combined_footsteps}
  \vspace{-1em}
\end{figure*}

\begin{figure*}[t]
  \centering
    \includegraphics[width=\linewidth]{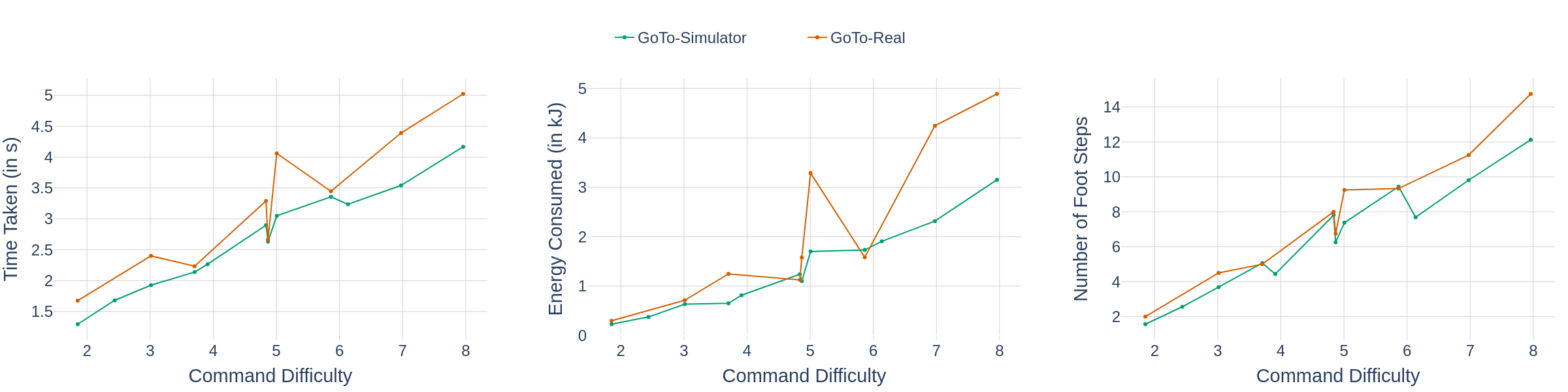}
    \caption{\small 
    Performance comparison between GoTo controller in simulation (green) and real-world deployment (orange) across increasing command difficulty. Metrics show time taken (left), energy consumption (middle), and footstep count (right) have slightly higher values for real-world deployment. However, the consistent trend patterns and minimal gap in the metrics validate successful sim-to-real transfer.}
  \label{fig:real_metrics}
  \vspace{-1em}
\end{figure*}

\subsection{Quantitative Simulation Results}

At the start of each simulation trial, (1) Digit's joints are initialized near the nominal standing pose with small random perturbations and (2) we sample an $\mathrm{SE}(2)$ goal by choosing a distance $r \in \{0.5,\,1.0,\,2.0,\,4.0\}\,\mathrm{m},$ an approach angle $\phi \in \{0,\,\tfrac{\pi}{4},\,\tfrac{\pi}{2},\,\tfrac{3\pi}{4}\}$ to place the target at $ \bigl(r\cos\phi,\;r\sin\phi\bigr),$ and a commanded final heading $\theta \in \{0,\,\tfrac{\pi}{4},\,\tfrac{\pi}{2},\,\tfrac{3\pi}{4}\}$ that the robot should adopt upon arrival. A trial is marked successful when the robot settles into a stable standing pose with both feet grounded, and its linear and angular velocities each fall below $0.1\,\mathrm{m/s}$ and $0.1\,\mathrm{rad/s}$, respectively. Our evaluation is based on running 16 trials for each $\mathrm{SE}(2)$ command, which have variance due to the randomly perturbed initial pose. We use the metrics defined in Section~\ref{sec:prob_form} to evaluate our GoTo controller against the baselines.

Table \ref{tab:overall} summarizes the performance metrics averaged across all trials for all five controllers, with values normalized relative to the Agility-2 controller (set to 1.0) to facilitate direct comparison. The GoTo controller outperforms all baselines across every metric, surpassing even the hand-tuned version of the manufacturer-provided Agility controller (Agility-2). It reduces energy consumption by 37\% (0.63±0.19), time-to-target by 44\% (0.56±0.09s), and footstep count by 33\% (0.67±0.15) while improving position accuracy by 40\% (0.6±0.37cm) and orientation accuracy by 58\% (0.42±0.31rad). Lower standard deviations across all metrics further demonstrate the superior reliability of our approach.

The hierarchical controller offers moderate improvements in efficiency metrics compared to FSM and Agility-3, but shows higher position errors. The FSM approach, with its rigid orient-then-move strategy, performs poorly across nearly all metrics, highlighting the limitations of phase-separated control on top of a lower-level velocity controller. The Agility-3 controller, despite being the default controller, underperforms significantly compared to the Agility-2 controller on all metrics, particularly with high variability in energy consumption, showcasing the sensitivity to design choices for model-based controllers. 

Figure \ref{fig:combined_footsteps} shows time, energy, and step count for each controller, grouped by command difficulty. In particular, the command difficulty of an $\mathrm{SE}(2)$ offset command is measured according to the time required for the Agility-2 controller to achieve the target. We see that the GoTo controller uniformly outperforms other controllers across all difficulties and metrics. The relative improvement of GoTo over the baselines also tends to increase with difficulty. This again shows the value in training a dedicated end-to-end controller for $\mathrm{SE}(2)$-target locomotion.

\subsection{Qualitative Footstep Results}

\begin{figure}[t]
  \centering


    \includegraphics[width=0.9\linewidth]{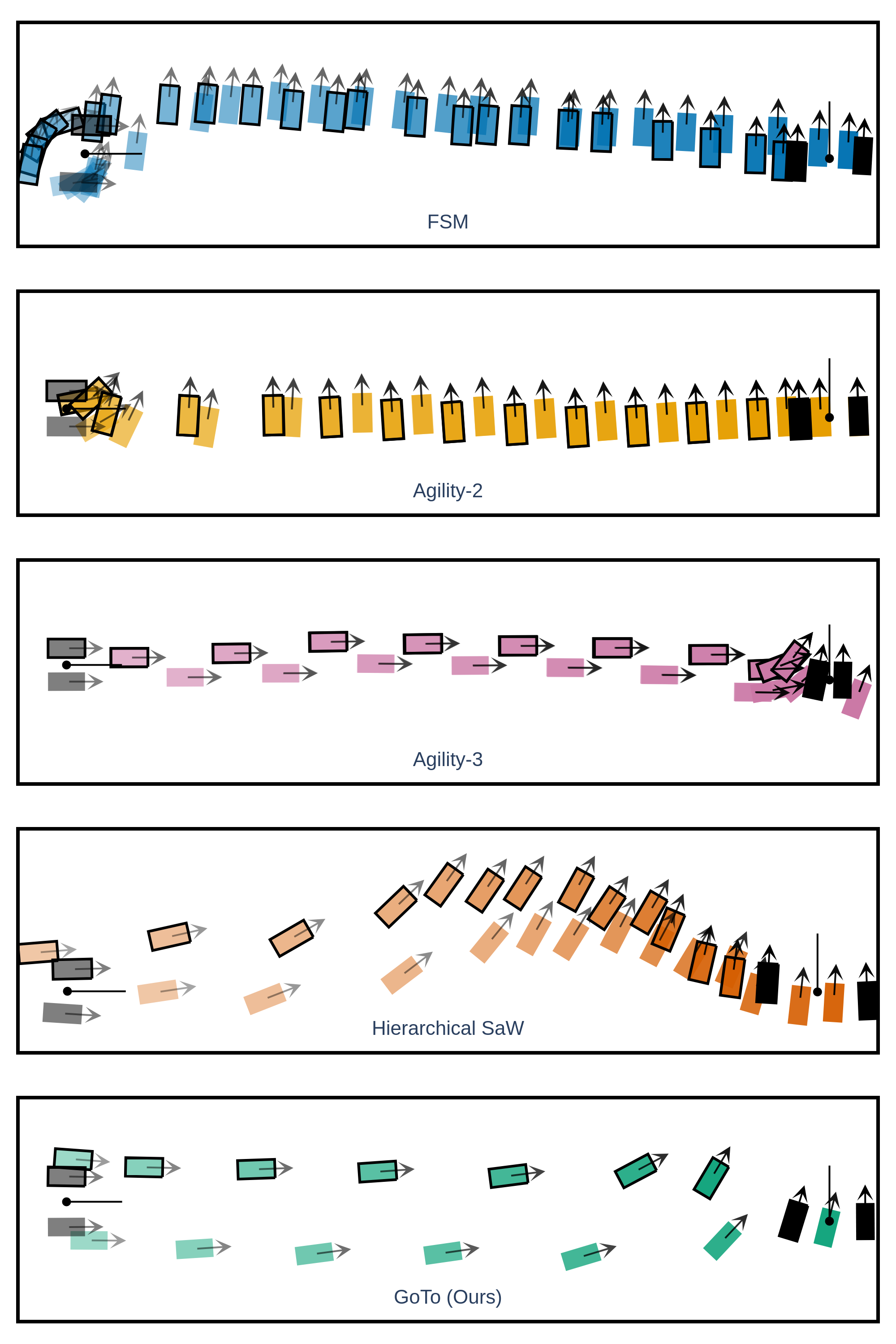}
    \caption{\small Footstep trajectories for the FSM, Agility-2, Agility-3, Hierarchical SaW, and GoTo controllers. The initial pose (origin, facing east) and final pose (northward at the 4m goal) are marked with black dots and arrows. Each footstep is shown as a semi-transparent colored rectangle: left feet have black borders, right feet do not. Arrowheads indicate foot orientation. Standing poses at the start and end are shown in black.}
    \vspace{-1em}

  \label{fig:footsteps}
\end{figure}

Figure \ref{fig:footsteps} shows the typical footstep patterns of the different controllers for an $\mathrm{SE}(2)$ target with a position delta of 4m and an orientation delta of 90deg. By design, we see the FSM employs a rigid, two-phase approach---first aligning to goal orientation, then executing sidestep marching. Similarly, the default Agility 1 controller follows this pattern with more consistent gait structure.  The hand-tuned Agility 2 controller improves efficiency by marching directly to the target position with structured footsteps, deferring orientation adjustments until reaching the destination. The Hierarchical controller takes a hybrid approach, beginning orientation optimization midway through its trajectory, with marching behavior intensifying once the orientation reward activates.

In contrast, the learned GoTo controller performs better both quantitatively and qualitatively. It uses fewer steps while producing natural movement patterns. Instead of treating rotation and translation as separate phases, GoTo generates deliberate, purposeful steps that seamlessly integrate directional changes into forward motion. This integration creates fluid, human-like locomotion where turning occurs organically during travel---eliminating the need for mechanical marching or discrete rotation phases. The result combines greater efficiency (fewer total steps) with more natural movement (continuous, coordinated locomotion), representing a clear advancement over the segmented or rigid behaviors exhibited by all baseline controllers.



\subsection{Real-World Results}

To validate the effectiveness of our approach beyond simulation, we deployed the trained GoTo controller on the physical Digit v3 humanoid platform. We evaluated the performance across 72 distinct short-range $\mathrm{SE}(2)$ navigation tasks sampled from a grid of distances \(\{0.5, 1.0, 2.0\}\;\mathrm{m}\) and orientation deltas \(\{0, \frac{\pi}{2}, \frac{3\pi}{4}\}\;\mathrm{rad}\). For each distance-orientation pair, we conducted 8 trials varying either \(\Delta x\) or \(\Delta y\) to ensure coverage of the target workspace.

The GoTo controller achieved a \textbf{90.23\% success rate} across all trials, compared to 100\% in simulation. The primary failure mode involved the robot losing balance and falling during execution of more challenging commands, particularly those combining longer distances with large orientation changes. These trials were excluded from our performance analysis to maintain consistent comparisons with simulation results. Figure~\ref{fig:real_metrics} shows the performance metrics plotted against command difficulty, analogous to our simulation analysis. The real-world trends closely mirror those observed in simulation, with the GoTo controller maintaining its efficiency advantages across all difficulty levels. 

Despite the challenges inherent in transferring policies to physical hardware—including sensing noise, actuation delays, and unmodeled dynamics—the controller produced stable, purposeful locomotion. These natural movements and close sim and real world benchmark numbers observed confirm that the constellation reward effectively promotes behaviors that transfer robustly from simulation to real-world execution without additional fine-tuning, highlighting the immediate applicability of our approach for practical deployment scenarios.

\section{CONCLUSION}

This work presents a novel approach to short-range humanoid locomotion by directly learning end-to-end policies optimized for $\mathrm{SE}(2)$ target reaching. Departing from traditional velocity-tracking strategies, we introduce the \textit{constellation reward}—a unified geometric objective that naturally couples translation and rotation without requiring finely tuned trade-offs. Across simulation and real-world deployments, our \textit{GoTo} controller consistently outperforms baselines, producing smoother trajectories, fewer footsteps, and significantly lower energy consumption.
While the policy demonstrates strong sim-to-real transfer, occasional failures under extreme conditions highlight opportunities for improving robustness. Future extensions could explore broader task objectives, such as incorporating whole-body pose goals and end-effector constraints, enabling seamless integration of locomotion and manipulation.
Altogether, these results highlight the value of task-specific reward design and end-to-end learning for practical humanoid control, offering a promising foundation for agile, goal-directed locomotion in real-world environments.






\bibliographystyle{IEEEtran}
\bibliography{references}

\end{document}